\documentclass[10pt,twocolumn,letterpaper]{article}

\usepackage[pagenumbers]{cvpr} 

\definecolor{cvprblue}{rgb}{0.21,0.49,0.74}
\usepackage[pagebackref,breaklinks,colorlinks,allcolors=cvprblue]{hyperref}

\def\paperID{10866} 
\def\confName{CVPR}
\def\confYear{2025}

\def\@fnsymbol#1{\ensuremath{\ifcase#1\or *\or \dagger\or \ddagger\or
   \mathsection\or \mathparagraph\or \|\or **\or \dagger\dagger
   \or \ddagger\ddagger \else\@ctrerr\fi}}

\title{Large Motion Video Autoencoding with Cross-modal Video VAE}
\author{Yazhou Xing\thanks{equal contribution}  \quad Yang Fei$^{*}$ \quad Yingqing He$^{*}$\thanks{corresponding authors} \quad Jingye Chen\quad Jiaxin Xie \\ \quad Xiaowei Chi \quad Qifeng Chen$^{\dagger}$
\\
The Hong Kong University of Science and Technology
}

\begin{document}
\maketitle
\begin{abstract}

Learning a robust video Variational Autoencoder (VAE) is essential for reducing video redundancy and facilitating efficient video generation. 
Directly applying image VAEs to individual frames in isolation can result in temporal inconsistencies and suboptimal compression rates due to a lack of temporal compression. 
Existing Video VAEs have begun to address temporal compression; however, they often suffer from inadequate reconstruction performance.
In this paper, we present a novel and powerful video autoencoder capable of high-fidelity video encoding. 
First, we observe that entangling spatial and temporal compression by merely extending the image VAE to a 3D VAE can introduce motion blur and detail distortion artifacts. 
Thus, we propose temporal-aware spatial compression to better encode and decode the spatial information. 
Additionally, we integrate a lightweight motion compression model for further temporal compression.
Second, we propose to leverage the textual information inherent in text-to-video datasets and incorporate text guidance into our model. 
This significantly enhances reconstruction quality, particularly in terms of detail preservation and temporal stability.
Third, we further improve the versatility of our model through joint training on both images and videos, which not only enhances reconstruction quality but also enables the model to perform both image and video autoencoding. 
Extensive evaluations against strong recent baselines demonstrate the superior performance of our method. The project website can be found at~\href{https://yzxing87.github.io/vae/}{https://yzxing87.github.io/vae/}. 

\end{abstract}    
\section{Introduction}
\label{sec:intro}

\begin{figure*}[t]
\centering
\includegraphics[width=1.0\textwidth]{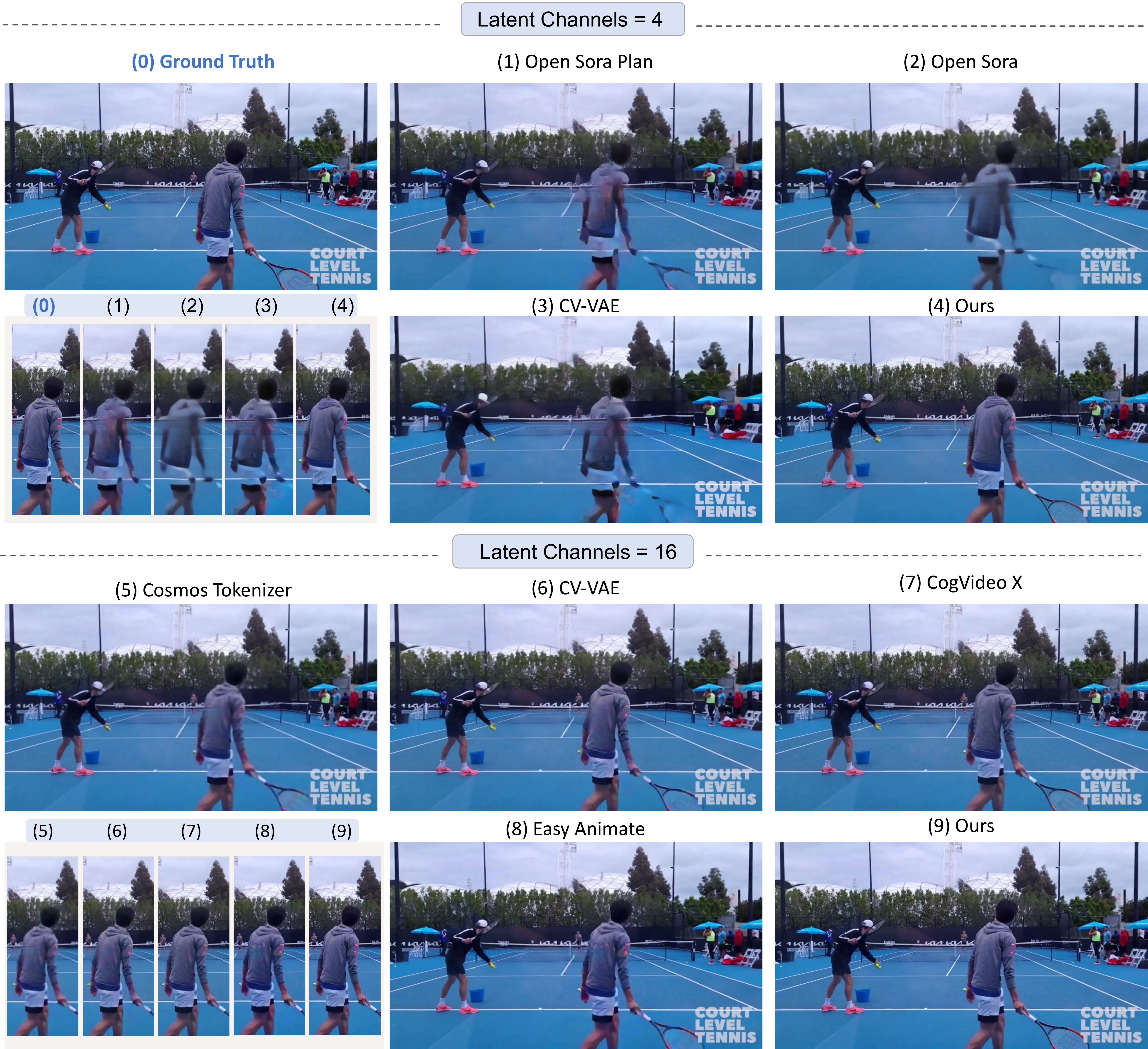}
\caption{
Our reconstruction results compared with a line of three recent strong baseline approaches. 
The ground truth frame is (0). Our model significantly outperforms previous methods, especially under large motion scenarios such as people doing sports.
}
\label{fig:teaser}
\vspace{-3mm}
\end{figure*}

Given the significant attention in the field of video generation, Latent Video Diffusion Models (LVDMs)~\cite{blattmann2023stable, blattmann2023align, he-lvdm, zhou2022magicvideo, he-videocrafter1} have emerged as a popular framework. They have been successfully applied to powerful text-to-video models such as Sora~\cite{videoworldsimulators2024}, VideoCrafter~\cite{he-videocrafter1, chen2024videocrafter2overcomingdatalimitations}, and CogVideoX~\cite{yang2024cogvideox}.
Different from directly generating video pixels, LVDMs generate latent video representations in a compact latent space. This is achieved by first training a Video VAE to encode videos into this latent space.
Thus, Video VAE, as a key and fundamental component of LVDMs, has attracted great attention recently.
An effective Video VAE can help to reduce the training costs of video diffusion models while improving the final quality of the generated videos.
Initially, a series of studies adopt the image VAE from Stable Diffusion~\cite{rombach2022high} for video generation tasks, including AnimateDiff~\cite{guoanimatediff}, MagicVideo~\cite{zhou2022magicvideo}, VideoCrafter1~\cite{he-videocrafter1}, and VideoCrafter2~\cite{chen2024videocrafter2overcomingdatalimitations}. 
However, directly adopting an image VAE and compressing video on a frame-by-frame basis leads to temporal flickering due to the lack of temporal correlation. Additionally, the information redundancy along the temporal dimension is not reduced, leading to low training efficiency for subsequent latent video diffusion models.
From the introduction of Sora, which compresses videos both temporally and spatially through a Video VAE, a series of studies have emerged that aim to replicate Sora and train their own Video VAEs, including Open Sora~\cite{opensora}, Open Sora Plan~\cite{pku_yuan_lab_and_tuzhan_ai_etc_2024_10948109}, CV-VAE~\cite{zhao2024cv}, CogVideoX~\cite{yang2024cogvideox}, EasyAnimate~\cite{xu2024easyanimatehighperformancelongvideo}, and Cosmos Tokenizer~\cite{cosmos_token}.
However, the performance of the current video VAE suffers from many problems, including motion ghost, low-level temporal flickering, blurring (faces, hands, edges, texts), and motion stuttering (lack of correct temporal transition).

In this work, we propose a novel cross-modal Video VAE with better spatial and temporal modeling ability in order to solve the aforementioned challenge problems and obtain a robust and high-quality Video VAE.
First, we examine different designs for spatial and temporal compression, including simultaneous spatial-temporal (ST) compression and sequential ST compression. 
We observed that simultaneous ST compression achieves better low-level temporal smoothness and texture stability, while sequential ST compression achieves better motion recovery, particularly in scenarios of large motion.
Thus, we propose a novel architecture that integrates the advantages of both methods and enables effective video detail and motion reconstruction.

Second, we observed that the normally used datasets for text-to-video generation contain text-video pairs. 
Also, during decoding, a text description exists as it serves as the input in the first stage, \textit{i.e.}, the video latent generation stage.
To this end, we integrate the text information into the encoding and decoding procedure and propose the first Cross-modal Video VAE.
We carefully study how text guidance can be integrated into the spatiotemporal backbone and the mechanism of spatial and temporal semantic guidance. 

In addition, our cross-modal video VAE supports image-video joint training.
To achieve this, we design our network with a fully spatiotemporal factorized architecture, and we feed image and video batches alternately to the network. 
During image batches, the data only forwards the spatial part of the network, with the temporal modules being skipped. During video batches, the video forwards both spatial and temporal modules. We also demonstrate that image joint training is crucial for training a video VAE.
In summary, our contributions are as follows:
\begin{itemize}
    \item We propose an effective and robust Video VAE, conduct extensive experiments, and achieve the state-of-the-art.
    \item We propose an optimal spatiotemporal modeling approach for Video VAE.
    \item We propose the first cross-modal video VAE that leverages the information from other modalities, i.e., text descriptions, to the best of our knowledge.
    \item Our video VAE is designed and trained to be versatile to conduct both image and video compression. 
\end{itemize}

\section{Related Work}
\label{sec:relat}

\paragraph{Video Variational Autoencoder} Video Variational Autoencoders (VAEs)~\cite{kingma2014auto} can be broadly categorized into discrete and continuous types. Discrete video VAEs compress videos into discrete tokens by learning a codebook for quantization and have achieved state-of-the-art performance in video reconstruction, as demonstrated by models like MAGVIT-v2~\cite{yu2023language}. However, these VAEs are not suitable for Latent Video Diffusion Models (LVDMs)~\cite{he-lvdm} due to the lack of necessary gradients for backpropagation, which hinders smooth optimization.

In contrast, continuous Video VAEs compress videos into continuous latent representations that are widely adopted in LVDMs. 
In earlier video generation studies, including Stable Video Diffusion~\cite{blattmann2023stable}, the Video VAE was directly adapted from the image VAE used in Stable Diffusion~\cite{rombach2022high}, achieving a compression ratio of $1 \times 8 \times 8$  by processing each frame independently. 
To further reduce the temporal redundancy, more recent studies~\cite{zhao2024cv,pku_yuan_lab_and_tuzhan_ai_etc_2024_10948109,opensora,xu2024easyanimatehighperformancelongvideo,yang2024cogvideox} have trained their VAEs to achieve a more efficient compression ratio of $4 \times 8 \times 8$. 

Despite these advancements, all of the aforementioned video VAEs struggle with accurately reconstructing videos with large motions due primarily to their limited ability to handle the temporal dimension effectively. A high-quality Video VAE that can robustly reconstruct videos with significant motion is critical in the LVDM pipeline, as it ensures efficient latent space compression, maintains temporal coherence and reduces computational overhead~\cite{metamoviegen}. Without a robust VAE, large motions in videos can lead to poor latent representations, negatively impacting the quality and overall performance of the LVDMs.

\paragraph{Latent Video Diffusion Models} 
Latent Video Diffusion Models (LVDMs) are widely used in foundational video generation models including Sora 
~\cite{videoworldsimulators2024}, OpenSora~\cite{opensora}, Open Sora Plan~\cite{pku_yuan_lab_and_tuzhan_ai_etc_2024_10948109}, VideoCrafter1~\cite{he-videocrafter1}, VideoCrafter2~\cite{chen2024videocrafter2overcomingdatalimitations}, Latte\cite{ma2024latte}, CogVideoX~\cite{yang2024cogvideox}, DynamiCrafter~\cite{xing2023dynamicrafter}, Vidu~\cite{bao2024vidu}, Hunyuan Video~\cite{kong2024hunyuanvideo}, controllable video generation~\cite{he-animate-a-story, follow-your-pose, follow-your-emoji}, and multimodal video generation models~\cite{he-seeing-and-hearing, he-llm-survey}.
%
The general pipeline for these LVDMs consists of two primary steps. First, the raw video is compressed into a latent space via a video Variational Autoencoder (VAE), significantly reducing computational complexity. In the second step, a diffusion model operates within this latent space, learning the desired transformations. The performance of LVDMs is critically dependent on video VAEs, as the quality of the generated video is heavily influenced by the latent space representation and the encoding-decoding capabilities of the VAE.

In image generation tasks, Stable Diffusion series~\cite{rombach2022high, podell2023sdxl, sd35} has excelled, largely due to its efficient VAE that reconstructs diverse image types with high fidelity. However, no existing VAE in video generation achieves comparable quality, particularly due to challenges in compressing the temporal dimension. This limitation hinders the performance of LVDMs, especially in high-motion scenarios.

\begin{figure*}[t]
\centering
\includegraphics[width=1.0\textwidth]{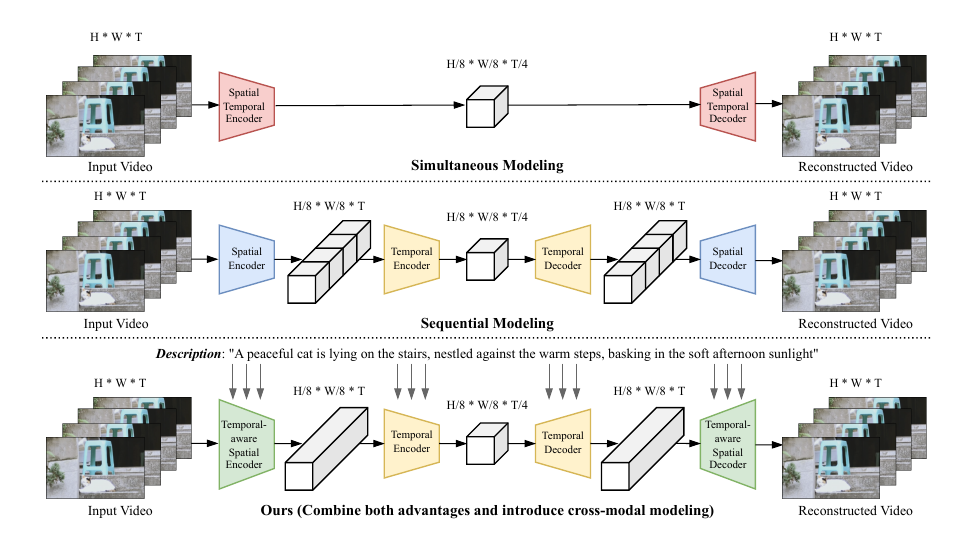}
\caption{Comparison of our optimal spatiotemporal modeling and the two other options. Simultaneous modeling is achieved by inflating pre-trained 2D spatial VAE to 3D VAE. Sequential modeling indicates first compressing the spatial dimension with a spatial encoder and then compressing the temporal information with a temporal encoder. 
We identify the issues of these two options and propose to combine both advantages and achieve a much better video reconstruction quality. 
Our VAE also benefits from cross-modality, i.e., text information. 
}
\label{fig:framework}
\vspace{-3mm}
\end{figure*}

\section{Method}

\subsection{Overview}
The video autoencoding problem can be defined as follows. Let $\mathbf{X} \in \mathbb{R}^{C \times T \times H \times W}$ represent a video or image tensor, where $C$, $T$, $H$, and $W$ denote the number of channels, frame(s), height, and width, respectively. We want to train an encoder $\mathcal{E}$ that compresses the input tensor $\mathbf{X}$ into a compact latent representation $\mathbf{Z} \in \mathbb{R}^{C' \times T' \times H' \times W'}$.
The learned compact latent $\mathbf{Z}$ can be further reconstructed back to RGB space with decoder $\mathcal{D}$:

\begin{equation}
  \mathbf{Z} = \mathcal{E}({\mathbf{X}}), \hat{\mathbf{X}} = \mathcal{D}(\mathbf{Z}). 
\end{equation}

Our goal is to design and learn such an autoencoder that can reduce the spatial and temporal dimension of video data in latent space and reconstruct the video with highly spatial and temporal fidelity, especially for large-motion scenarios.

We first examine two inherited video VAE designs from the pre-trained Stable Diffusion model. We then combine the best of two designs and propose our spatiotemporal modeling that can reconstruct high-dynamic contents with fine details. 
We then investigate the text-conditioned video autoencoding and propose an effective text-guided video VAE architecture. 
Moreover, we propose a joint image and video compression training method, that enables text-aided joint image and video autoencoding. Our method does not rely on causal convolution as adopted by prior works. Finally, we carefully study the effects of different loss functions on the reconstruction performance and present the state-of-the-art video VAE architecture.

\subsection{Optimal Spatiotemporal Modeling}
Designing a video VAE that is inherited from a pre-trained 2D spatial VAE is a good practice to leverage the spatial compression prior. There are typically two options to inflate a 2D spatial VAE to its 3D video counterpart. 

\paragraph{Option 1: Simultaneous Spatiotemporal Compression} One common way to inherit the weight from pre-trained 2D VAE is to inflate the 2D spatial blocks to 3D temporal blocks and simultaneously do the spatial and temporal compression. We first examine this design. Specifically, we replace the 2D convolution in SD VAE with 3D convolution of kernel size (1,3,3), whose weights are initialized from the 2D convolution. Then we add an additional temporal convolution layer with kernel size (3,3,3) to learn spatiotemporal patterns. In this middle block of the inflated VAE, we inflate the 2D attention to 3D attention and we also include a temporal attention to capture both the spatial and temporal information. We keep other components unchanged to maximumly leverage the learned prior of SD VAE. 

\paragraph{Option 2: Sequential Spatiotemporal Compression}
Another reasonable way to cooperate the SD VAE to video VAE is to keep the SD VAE unchanged: first utilize the SD VAE to compress the input video frame-by-frame, and then learn a temporal autoencoding process to further compress the temporal redundancy, as shown in Fig.~\ref{fig:framework}. Specifically, we adopt a lightweight temporal autoencoder for temporal compression. The encoder consists of one convolutional layer to process the input, and two or three 3D ResNet blocks with convolutional downsampling layers to compress the temporal redundancy. Notably, we design the decoder to be asymmetric as the encoder, i.e., there will be two 3D ResNet blocks following each upsampling layer in the decoder. Through this asymmetric design, our decoder can potentially gain some hallucination ability beyond the reconstruction. 

\begin{figure*}[t]
\centering
\includegraphics[width=0.8\textwidth]{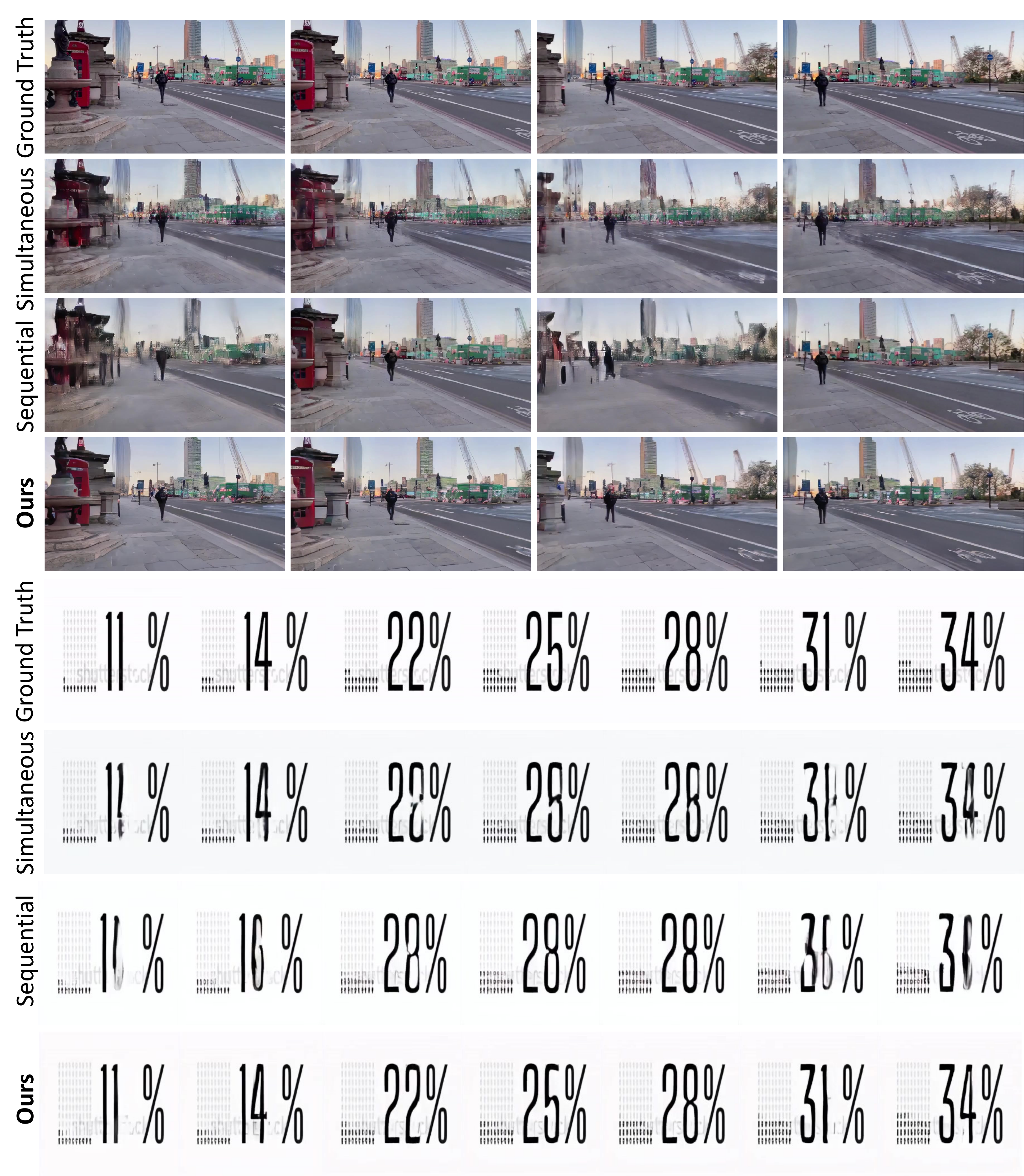}
\caption{Comparisons among simultaneous spatiotemporal modeling, sequential spatiotemporal modeling and our proposed solution. 
}
\label{fig:modeling}
\vspace{-3mm}
\end{figure*}

Surprisingly, we find this sequential spatiotemporal design can better compress and recover the dynamic of the input video than option 1, but is not good at recovering spatial details, which is proved by consistent improvement under large-motion video autoencoding as shown in Fig.~\ref{fig:modeling}. 

\begin{figure}[t]
\centering
\includegraphics[width=0.5\textwidth]{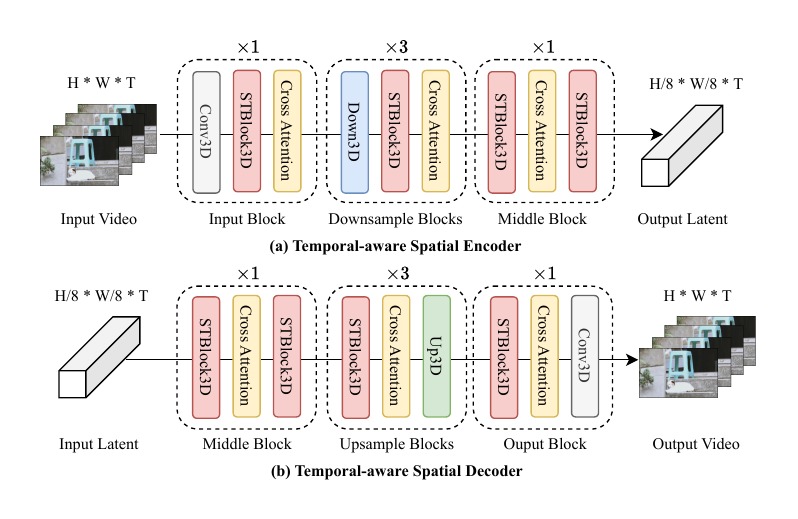}
\caption{The architecture of our temporal-aware spatial autoencoder. We expand the 2D convolution of SD VAE~\cite{rombach2022high} to 3D convolution and append one additional 3D convolution as temporal convolution after the expanded 3D convolution, which forms the STBlock3D. We also inject the cross-attention layers for cross-modal learning with textual conditions. }
\label{fig:2+1D}
\vspace{-3mm}
\end{figure}

\paragraph{Our Solution} We find simultaneous spatiotemporal compression leads to better detail-recovering capability, and the sequential spatiotemporal compression will exceed at motion-recovering ability. Thus, we propose to combine the best of two worlds, and introduce the two-stage spatiotemporal modeling for video VAE. As the first stage, we inflate the 2D convolution to 3D convolution with kernel size (1,3,3), and similarly to option 1, we add additional temporal convolution layers through 3D convolution. We denote our first-stage model as a temporal-aware spatial autoencoder. Different from option 1, we only compress the spatial information and do not compress the temporal information at the first stage, but introduce another temporal encoder to further encode the temporal dimensions, which serves as the second stage compression. We follow the same design of option 2 for our temporal encoder and decoder. After that, we decode the reconstructed latent of the second stage to the RGB space, with the inflated 3D decoder. We jointly train the inflated 3D VAE and the temporal autoencoder. The main idea is illustrated in Fig.~\ref{fig:framework} and Fig.~\ref{fig:2+1D}. 

\paragraph{Formulation}
Recall \( \mathbf{X} \in \mathbb{R}^{C \times T \times H \times W} \) represent a video, where \( C \), \( T \), \( H \), and \( W \) denote the number of channels, frames, height, and width, respectively. The \( i \)-th frame of the video is denoted as \( \mathbf{x}_i \in \mathbb{R}^{C \times H \times W} \). The temporal-aware spatial encoder encodes \( \mathbf{X} \) into a latent representation \( \mathbf{Z}_1 \in \mathbb{R}^{c \times T \times h \times w} \), where \( c \) is the number of latent channels, and \( h = \frac{H}{8} \), \( w = \frac{W}{8} \), as formulated by:

\begin{equation}
    \mathbf{Z}_1 = \mathcal{E}_1(\mathbf{X}).
\end{equation}

Next, the temporal autoencoder encodes \( \mathbf{Z}_1 \) into \( \mathbf{Z}_2 \in \mathbb{R}^{c' \times t \times h \times w} \), where \( c' \) is the number of latent channels for \( \mathbf{Z}_2 \) and \( t = \frac{T}{4} \), as given by:

\begin{equation}
    \mathbf{Z}_2 = \mathcal{E}_2(\mathbf{Z}_1).
\end{equation}

Reconstruction is achieved by decoding \( \mathbf{Z}_2 \) back into the original video space, \( \hat{\mathbf{X}} \in \mathbb{R}^{C \times T \times H \times W} \), through the following inverse process:

\begin{equation}
    \hat{\mathbf{X}} = \mathcal{D}_1(\mathcal{D}_2(\mathbf{Z}_2)) = \mathcal{D}_1(\mathbf{Z}_1).
\end{equation}

\subsection{Cross-modal Modeling}
Since textual information is a native component for text-to-video generation datasets, we examine if the textual information can improve the autoencoding process of the model. To achieve that, we split the feature maps into patches as tokens after each ResNet block in the encoder and decoder, and compute the cross attention by taking visual tokens as query (Q) and value (V), the text embeddings as key (K). 

We try to keep the patch size trackable for each layer. Specifically, we use patch size to 8$\times$8, 4$\times$4, 2$\times$2, and 1$\times$1 for each layer in the temporal-aware spatial autoencoder respectively. We directly use each pixel as one patch in the temporal autoencoder.  We adopt LayerNorm as the normalization function. We use Flan-T5~\cite{t5} as the text embedder. 
A projection convolution is applied to the result, which is then added to the input via a residual connection.

\subsection{Joint Image and Video Compression}
In contrast to existing architectures such as MagVitV2~\cite{yu2023language}, OD-VAE~\cite{chen2024odvaeomnidimensionalvideocompressor}, and OPS-VAE~\cite{opensora}, which use Causalconv3D layers, we rely primarily on standard Conv3D layer.

A notable feature of our architecture is the ability to mask out the temporal autoencoder, allowing the first-stage model to operate as a standalone image compressor. 
During training, our model is flexible to take both image and video as input: when the current batch is composed of images, we will disable the temporal convolution and temporal attention layers, as well as the temporal autoencoder. We train our model on both the image dataset and video dataset to let the model learn the image and video compression ability simultaneously. Besides, training on more high-quality images can also help improve the video autoencoding performance. We quantitatively evaluate the performance of our joint image and video compression in Table~\ref{tab:main}.

\subsection{Loss Functions}
We use the reconstruction loss, the KL divergence loss, and the video adversarial loss (3D GAN loss) to optimize our model. 
The reconstruction loss, $\mathcal{L}_{\text{recon}}$, ensures that the generated frames are perceptually and structurally similar to the input frames. It combines a pixel-wise error term with a perceptual loss, weighted by a hyperparameter.
The KL divergence loss, $\mathcal{L}_{\text{KL}}$, regularizes the latent space by encouraging it to conform to a prior distribution, ensuring smoothness and continuity in the learned latent representations. Given the hierarchical structure of our latent space, we only regularize the innermost latent $\mathbf{Z_2}$, with dimensions $\frac{T}{4} \times \frac{H}{8} \times \frac{W}{8}$, where $T$, $H$, and $W$ represent the temporal, height, and width dimensions, respectively. The 3D GAN loss, $\mathcal{L}_{\text{GAN}}$, is introduced to enhance the realism of the generated video sequences, leveraging a discriminator to distinguish between real and generated sequences.
The total loss function is expressed as:

\begin{equation}
    \mathcal{L}_{\text{total}} = \mathcal{L}_{\text{recon}} + \lambda_{\text{KL}} \mathcal{L}_{\text{KL}} + \lambda_{\text{GAN}} \mathcal{L}_{\text{GAN}}.
\end{equation}

\begin{table*}[ht]
    \centering
    \resizebox{\textwidth}{!}{%
        \setlength\tabcolsep{5pt} 
        \renewcommand{\arraystretch}{1.2} 
        \begin{tabular}{lccccccccccc}
            \toprule
            \textbf{Model} & \textbf{Downsample Factor} & \textbf{\#Channels} & \multicolumn{3}{c}{\textbf{WebVid Test Set~\cite{bain2021frozen}}} & \multicolumn{3}{c}{\textbf{Inter4K Test Set~\cite{inter4K}}} & \multicolumn{3}{c}{\textbf{Large Motion Test Set}}\\
            \cmidrule(lr){4-6} \cmidrule(lr){7-9} \cmidrule(lr){10-12}
            & & & \textbf{PSNR ($\uparrow$)} & \textbf{SSIM ($\uparrow$)} & \textbf{LPIPS ($\downarrow$)} & \textbf{PSNR ($\uparrow$)} & \textbf{SSIM ($\uparrow$)} & \textbf{LPIPS ($\downarrow$)} & \textbf{PSNR ($\uparrow$)} & \textbf{SSIM ($\uparrow$)} & \textbf{LPIPS ($\downarrow$)} \\
            \midrule
            Open-Sora-Plan (OD VAE~\cite{chen2024odvaeomnidimensionalvideocompressor}) & 4x8x8 & 4 & 29.1646 & 0.8334 & 0.0789 & 28.6690 & 0.8381 & 0.0906 & \underline{\underline{27.5697}} & \underline{\underline{0.8045}} & 0.1065 \\
            Open-Sora (OPS VAE~\cite{opensora})& 4x8x8 & 4 & 29.3753 & 0.8284 & 0.1240 & \textbf{29.2721} & 0.8431 & 0.1316 & \textbf{27.7586} & 0.8032 & 0.1540 \\
            CV-VAE~\cite{zhao2024cv}  & 4x8x8 & 4 & 28.6795 & 0.8154 & 0.1072 & 27.7437 & 0.8124 & 0.1284 & 26.9456 & 0.7849 & 0.1411 \\
            \textbf{Video VAE w/o Joint Training (Ours)} & 4x8x8 & 4 & \underline{30.2091} & \underline{0.8656} & \underline{\underline{0.0566}} & 28.9048 & \underline{0.8543} & \underline{\underline{0.0688}} & 27.3917 & \underline{0.8078} & \underline{\underline{0.0867}}     \\
            \textbf{Video VAE (Ours)} & 4x8x8 & 4 & \textbf{30.3140} & \textbf{0.8676} & \textbf{0.0538} & \underline{\underline{28.9227}} & \textbf{0.8565} & \textbf{0.0665} & \underline{27.6236} & \textbf{0.8136} & \textbf{0.0841} \\
            \textbf{Cross-Modal VAE (Ours)} & 4x8x8 & 4 & \underline{\underline{30.1110}} & \underline{\underline{0.8608}} & \underline{0.0544} & \underline{29.0357} & \underline{\underline{0.8510}} & \underline{0.0678} & 27.1754 & 0.7999 & \underline{0.0846} \\
            \midrule
            Cosmos-Tokenizer~\cite{cosmos_token} & 4x8x8 & 16 & 31.2545 & 0.8861 & 0.1030 & 31.2002 & 0.8957 & 0.1071 & 30.1619 & 0.8675 & 0.1194 \\
            CogVideoX-VAE~\cite{yang2024cogvideox} & 4x8x8 & 16 & 32.8940 & 0.9208 & 0.0504 & 32.5122 & 0.9229 & 0.0532 & 31.0906 & 0.8978 & 0.0685 \\
            EasyAnimate-VAE~\cite{xu2024easyanimatehighperformancelongvideo} & 4x8x8 & 16 & 32.1233 & 0.9085 & 0.0405 & 31.5066 & 0.9048 & 0.0572 & 30.5213 & 0.8846 & 0.0598 \\
            CV-VAE~\cite{zhao2024cv} & 4x8x8 & 16 & 32.2766 & 0.9080 & 0.0546 & 31.6129 & 0.9060 & 0.0642 & 30.7136 & 0.8868 & 0.0726 \\
            \textbf{Video VAE w/o Joint Training (Ours)} & 4x8x8 & 16 & \underline{\underline{33.8844}} & \underline{\underline{0.9334}} & \underline{\underline{0.0344}} & \underline{\underline{32.9416}} & \underline{\underline{0.9297}} & \underline{\underline{0.0409}} & \underline{\underline{31.8471}} & \underline{\underline{0.9073}} & \underline{\underline{0.0499}} \\
            \textbf{Video VAE (Ours)} & 4x8x8 & 16 & \underline{34.1558} & \underline{0.9362} & \textbf{0.0271} & \underline{33.3184} & \underline{0.9328} & \textbf{0.0316} & \underline{32.1503} & \textbf{0.9122} & \textbf{0.0409} \\
            \textbf{Cross-Modal VAE (Ours)} & 4x8x8 & 16 & \textbf{34.5022} & \textbf{0.9365} & \underline{0.0323} & \textbf{33.5687} & \textbf{0.9347} & \underline{0.0379} & \textbf{32.2387} & \underline{0.9117} & \underline{0.0481} \\
            \bottomrule
        \end{tabular}
    }
    \caption{Quantitative comparison with state-of-the-art methods.}
    \label{tab:main}
\end{table*}

\begin{table}[ht]
    \centering
    \setlength\tabcolsep{4pt} 
    \renewcommand{\arraystretch}{1} 
    \begin{tabular}{lcccc}
        \toprule
        \textbf{Model} & \textbf{\# Ch} & \textbf{PSNR ($\uparrow$)} & \textbf{SSIM ($\uparrow$)} & \textbf{LPIPS ($\downarrow$)} \\
        \midrule
        SD1.4~\cite{blattmann2023stable} & 4 & 30.2199 & 0.8974 & 0.0440 \\
        \textbf{Ours w/o JT$^*$} & 4 & 15.1001 & 0.5561 & 0.4339 \\
        \textbf{Ours} & 4 & \textbf{30.8650} & \textbf{0.9042} & \textbf{0.0397} \\
        \midrule
        SD3.5~\cite{sd35}  & 16 & \textbf{36.5208} & \textbf{0.9646} & \textbf{0.0116} \\
        \textbf{Ours w/o JT$^*$} & 16 & 9.2603 & 0.2770 & 0.6802 \\
        \textbf{Ours} & 16 & \underline{35.3437} & \underline{0.9590} & \underline{0.0167} \\
        \bottomrule
    \end{tabular}
    \caption{JT$^*$ means joint training. We evaluate image reconstruction performance w/ or w/o our joint image-video training strategy.}
    \label{tab:ablation_joint}
    \vspace{-3mm}
\end{table}

\section{Experiments}

\subsection{Experimental Setup}
\paragraph{Datasets} 
We conduct experiments on three datasets: the public Panda2M~\cite{chen2024panda} and MMTrailer~\cite{chi2024mmtrailmultimodaltrailervideo} datasets, and a private text-video dataset with over 6M pairs.
To evaluate reconstruction performance, we use three test sets: the WebVid test set, the Inter4K test set (similar to~\cite{zhao2024cv}), and a large motion test set. The WebVid test set contains 1,000 256x256, 16-frame videos from the WebVid dataset~\cite{bain2021frozen}. The Inter4K test set consists of 500 640x864, 16-frame videos from the Inter4K dataset~\cite{inter4K}.
To assess the model's ability to handle challenging motion patterns, we introduce a large motion test set. This set includes 80 videos from WebVid and 20 from Inter4K, manually selected for their complex motion dynamics.

\paragraph{Implementation Details}
We initialize our 4-channel and 16-channel latent Video VAEs from SD-1.4~\cite{rombach2022high} and SD-3.5~\cite{sd35}, respectively. For both models, we enable the video GAN loss after 50K warmup steps.
We initially train the 4-channel and 16-channel latent Video VAEs for 230K and 310K steps, respectively. Subsequently, we conduct joint image-video training, using an 8:2 video-to-image ratio to balance video and image reconstruction. For each training step, we sample 16 videos from Panda2M and our private text-video dataset, concatenating their frames into a single image batch. By masking the temporal dimension and bypassing the temporal autoencoder, we treat these images as independent static frames, allowing the model to learn from both temporal and spatial information.
The 4-channel and 16-channel latent Video VAEs undergo additional joint training for 100K and 185K steps, respectively.
For the cross-modal VAE, both models are initialized with their pre-trained weights. We train them on video-text pairs for 160K steps, enabling the model to learn the alignment between visual and textual modalities.

\subsection{Comparison with State-of-the-arts}

We compare our proposed Video VAE models with the state-of-the-art video compression models: Open-Sora-Plan~\cite{pku_yuan_lab_and_tuzhan_ai_etc_2024_10948109}, Open-Sora~\cite{opensora}, CV-VAE~\cite{zhao2024cv} on 4-channel latent models, and Cosmos-Tokenizer~\cite{cosmos_token}, CogVideoX~\cite{yang2024cogvideox}, EasyAnimate~\cite{xu2024easyanimatehighperformancelongvideo}, CV-VAE~\cite{zhao2024cv} on 16-channel models. 

\paragraph{Quantitative Evaluation}
We use PSNR, SSIM, and LPIPS~\cite{lpips} to quantitatively measure the quality of the reconstructed videos. We compare our method with baselines on our three test sets, as listed in Table~\ref{tab:main}. 
Among these, our 4-channel latent Video VAE demonstrates superior performance across most datasets and metrics. 
Specifically, our model achieves the best reconstruction quality on the WebVid test set, shown as more than 1dB improvements over baselines and a significant improvement on the LPIPS metrics, which indicates our reconstruction is both with high-fidelity and better perceptual quality. A similar conclusion can be made on the Inter4K test set. On the Large-Motion test set, our model maintains strong performance with a significant SSIM and LPIPS improvement, showcasing its robustness in handling complex motion scenarios.

For models with 16-channel latent space, our model consistently outperforms these baselines across all test sets. 
For example, on the WebVid test set, our model achieves more than 2dB in terms of PSNR, significantly higher than Cosmos-Tokenizer and CogVideoX. Moreover, our model achieves the best SSIM and LPIPS, demonstrating substantial improvements in both fidelity and perceptual quality.

In summary, our Video VAE models consistently outperform existing baselines across all test sets and metrics, highlighting their effectiveness in both low-channel (4-channel latent) and high-channel (16-channel latent) configurations.

\begin{figure}[t]
\centering
\includegraphics[width=0.48\textwidth]{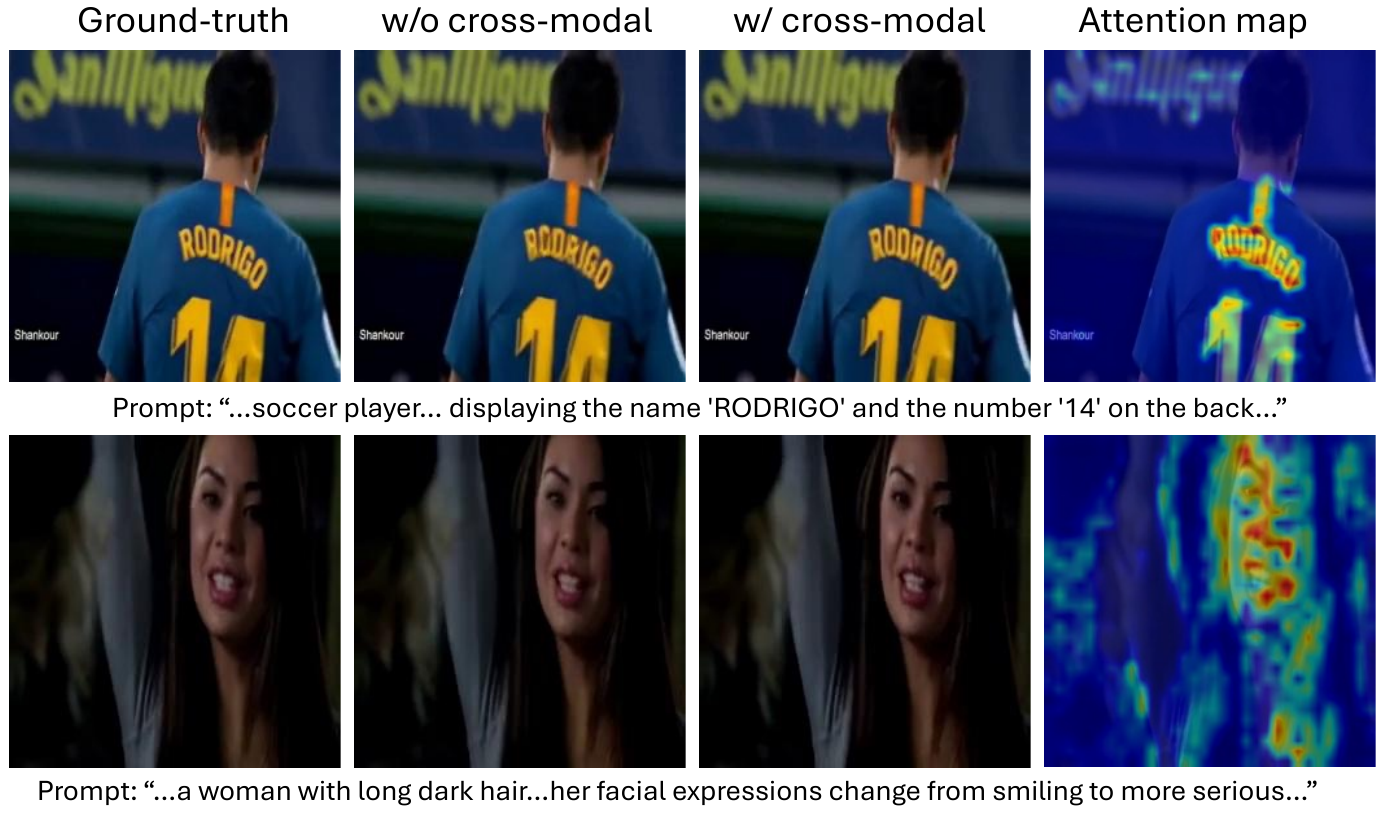}
\caption{The effectiveness of the cross-modal learning for our video VAE. The introduction of textural information improves the detail recovery. We visualize the learned attention map using keywords of the input prompts. 
}
\label{fig:cross}
\vspace{-3mm}
\end{figure}

\paragraph{Qualitative Evaluation}
We provide qualitative comparisons with the baselines in Fig.~\ref{fig:teaser}. Our method demonstrates significantly improved motion recovery, greatly reducing ghosting artifacts even in rapid motion scenarios. In contrast, Open-Sora-Plan and CV-VAE struggle to reconstruct fast-moving objects, leading to ghosting artifacts. Additionally, Open-Sora VAE introduces color reconstruction errors, as seen in the clothing of the moving figure. Increasing the latent channels to 16 improves motion reconstruction across all baselines, but noticeable detail errors remain. Our 16-channel model further mitigates these errors, resulting in more accurate detail reconstruction.
We further compare the reconstruction results with and without the cross-modal training, as shown in Fig.~\ref{fig:cross}.

\subsection{Ablation Study}

\paragraph{Joint Training}
We evaluate the effectiveness of our image-video joint training by comparing the performance of our 4-channel latent and 16-channel latent Video VAEs with the video-only training VAE, as well as the image VAE,  SD 1.4 and SD 3.5, respectively. The results are shown in Table~\ref{tab:main} and Table~\ref{tab:ablation_joint}. The video reconstruction comparison is conducted on the three benchmark datasets. The image reconstruction comparison is conducted on a set of 500 images with a resolution of 480x864, randomly sampled from a UHD-4K video dataset. 
During inference, we mask out the temporal autoencoder and the temporal part of the temporal-aware spatial autoencoder, ensuring that the models process the images without considering temporal information, effectively treating them as independent images.

The joint training can further boost the performance of video reconstruction, which is consistent in both the 4-channel and 16-channel experiments. 
For the image reconstruction, our 4-channel latent Video VAE slightly outperforms SD1.4, and also improves on SSIM and LPIPS, indicating better perceptual quality.

For the 16-channel VAE, while our model achieves competitive results in terms of PSNR, it falls slightly short of SD3.5. However, our model still demonstrates strong performance in terms of SSIM and LPIPS, suggesting that our joint training approach maintains high perceptual quality despite the slight drop in PSNR.

\begin{figure}[t]
\centering
\includegraphics[width=0.5\textwidth]{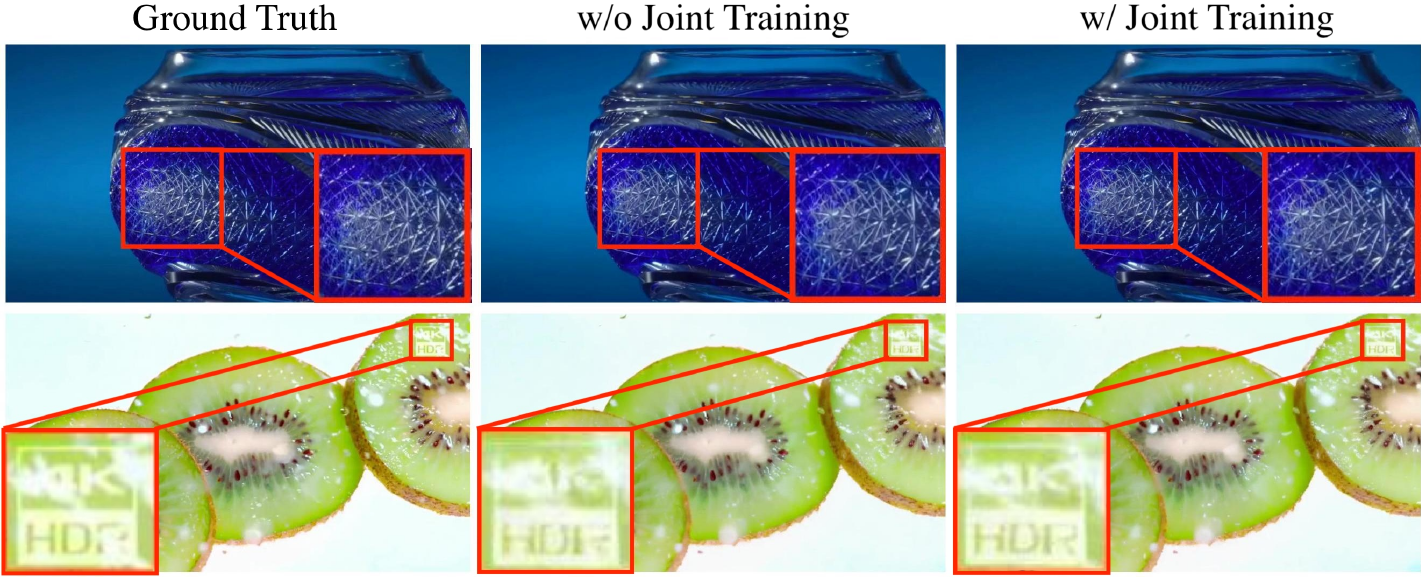}
\caption{The effectiveness of joint image and video training.   
}
\label{fig:joint}
\vspace{-5mm}
\end{figure}

We further show the visual effectiveness of the joint image and video training in Fig~\ref{fig:joint}. Overall, these results demonstrate that our joint image-video training strategy allows the model to retain strong image reconstruction capabilities while simultaneously learning to handle video data.

\begin{table}[ht]
    \centering
    \setlength\tabcolsep{2pt} 
    \renewcommand{\arraystretch}{1.2} 
    \begin{tabular}{lcccc}
        \toprule
        \textbf{Model} & \textbf{PSNR ($\uparrow$)} & \textbf{SSIM ($\uparrow$)} & \textbf{LPIPS ($\downarrow$)} \\
        \midrule 
        \textbf{Simultaneous } & \underline{24.0593} & \textbf{0.7315} & \underline{0.1293} \\
        \textbf{Sequential} & 23.3681 & 0.6917 & 0.1481 \\
        \textbf{Ours} & \textbf{24.6722} & \underline{0.7234} & \textbf{0.1162} \\
        \bottomrule
    \end{tabular}
    \caption{Ablation study comparing simultaneous modeling, sequential modeling, and ours on the large-motion test set.}
    \label{tab:ablation_architecture}
    \vspace{-5mm}
\end{table}

\paragraph{Architecture Variants}
We evaluate the effectiveness of different spatiotemporal compression strategies, including simultaneous spatiotemporal compression, sequential spatiotemporal compression, and our proposed solution. These architecture variants are tested on the Large-Motion Test Set to determine which model handles challenging scenarios most effectively, as shown in Table~\ref{tab:ablation_architecture}. 

\begin{table}[ht]
    \centering
    \setlength\tabcolsep{2pt} 
    \renewcommand{\arraystretch}{1} 
    \begin{tabular}{lcccc}
        \toprule
        \textbf{Model / Kernel Size} & \textbf{PSNR ($\uparrow$)} & \textbf{SSIM ($\uparrow$)} & \textbf{LPIPS ($\downarrow$)}  \\
        \midrule
        \textbf{Image GAN Loss} &  \underline{31.9133} & \underline{0.9071} & \underline{0.0436}  \\
        \textbf{Video GAN Loss} & \textbf{32.0262} & \textbf{0.9089} & \textbf{0.0426} \\
        \midrule
        \textbf{TemporalConv(3, 1, 1)} & 30.3332 & 0.8898 & 0.0489 \\
        \textbf{TemporalConv(5, 1, 1)} & 30.8745 & 0.9004 & 0.0475  \\
        \textbf{TemporalConv(7, 1, 1)} & 31.2922 & \underline{0.9025} & 0.0458  \\
        \textbf{TemporalConv(5, 3, 3)} & \underline{31.3516} & 0.9011 & \underline{0.0437}  \\
        \textbf{TemporalConv(7, 3, 3)} & \textbf{31.7444} & \textbf{0.9074} & \textbf{0.0436}  \\
        \bottomrule
    \end{tabular}
    \caption{Ablation study comparing temporal-aware spatial autoencoder with image/video GAN loss, and different kernel sizes.}
    \label{tab:ablation_combined}
    \vspace{-5mm}
\end{table}

\paragraph{Component Ablation}
We perform ablation studies on several key components of our model. First, we investigate the impact of the kernel size in the temporal convolutional layer of temporal-aware spatial autoencoder. The results of this study are shown in Table \ref{tab:ablation_combined}. Additionally, we explore the significance of the loss function by comparing the performance of temporal-aware spatial autoencoder trained with either the raw image GAN loss or the video GAN loss, with the results also presented in Table \ref{tab:ablation_combined}. These ablations are conducted on a validation set comprising 98 videos, each with a resolution of 256x256 pixels and a length of 16 frames, sourced from the MMTrailer dataset.

\section{Conclusion}
We propose a novel video variational autoencoder (VAE) to address high-fidelity video autoencoding and compression, especially for videos with large motion. 
Our approach extends pre-trained image VAEs to the video domain by decoupling spatial and temporal compression, mitigating motion blur and detail loss. 
We design a temporal-aware spatial encoder and a lightweight motion compression model to enhance motion modeling, temporal consistency, and detail preservation. 
To improve reconstruction quality and versatility, we leverage detailed captions and employ joint image-video training. Extensive experiments on challenging datasets demonstrate superior performance over state-of-the-art baselines. 
Our model sets a new standard for video compression by efficiently handling spatiotemporal compression while benefiting from cross-modal learning and joint training.
{
    \small
    \bibliographystyle{ieeenat_fullname}
    \bibliography{main}
}


\end{document}